\documentclass[11pt]{article}
\usepackage{epsfig}
\usepackage{apacite}
\usepackage{fullpage}

\newcommand{\epsfscaledbox}[2]{\centerline{\psfig{figure=#1,width=#2}}}
\newcommand{\todo}[1]{}
\newcommand{\boone}{\mbox{BO-}1}
\newcommand{\booone}{\mbox{BO-o}1}
\newcommand{\MLEone}{\mbox{MLE-}1}
\newcommand{\MLEoone}{\mbox{MLE-o}1}
\newcommand{\rand}{\mbox{RAND}}
\newcommand{\conf}{P_{\smrm{C}}}
\newcommand{\confarg}{P_{\smrm{C}}(w_1' | w_1)}
\newcommand{\p}{p(w_2)}
\newcommand{\q}{p'(w_2)}
\newcommand{\Lone}{L}
\newcommand{\JS}{J}
\newcommand{\smrm}[1]{\mbox{\scriptsize #1}}

\title{Similarity-Based Models of Word Cooccurrence Probabilities}

\author{\begin{tabular}{@{}p{11em}@{\hspace{1em}}p{11em}@{\hspace{1em}}p{11em}@{}}
\parbox[t]{11em}{\begin{center}
Ido Dagan \\
{\small\tt dagan@macs.biu.ac.il} \\
Dept.~of Mathematics and Computer Science \\
Bar Ilan University \\
Ramat Gan 52900, Israel\end{center}} &
\parbox[t]{11em}{\begin{center}Lillian Lee \\
{\small\tt llee@cs.cornell.edu } \\
Department of Computer Science \\
Cornell University \\
Ithaca, NY 14853, USA\end{center}} & 
\parbox[t]{11em}{\begin{center}Fernando C.~N.~Pereira \\
{\small\tt pereira@research.att.com} \\
AT\&T Labs -- Research \\
180 Park Ave. \\
Florham Park, NJ 07932, USA \end{center}}
\end{tabular}
}

\begin{document}
\maketitle
\begin{abstract}
In many applications of natural language processing (NLP) it
is necessary to determine the likelihood of a given word combination.
For example, a speech recognizer may need to determine which of the
two word combinations ``eat a peach'' and ``eat a beach'' is more
likely.  Statistical NLP methods determine the likelihood of a word
combination from its frequency in a training corpus.  However, the
nature of language is such that many word combinations are infrequent
and do not occur in any given corpus.  In this work we propose a
method for estimating the probability of such previously unseen word
combinations using available information on ``most similar'' words.

We describe probabilistic word association models based on
distributional word similarity, and apply them to two tasks, language
modeling and pseudo-word disambiguation.  In the language modeling
task, a similarity-based model is used to improve probability
estimates for unseen bigrams in a back-off language model.  The
similarity-based method yields a 20\% perplexity improvement in the
prediction of unseen bigrams and statistically significant reductions
in speech-recognition error.

We also compare four similarity-based estimation methods against
back-off and maximum-likelihood estimation methods on a pseudo-word
sense disambiguation task in which we controlled for both unigram and
bigram frequency to avoid giving too much weight to easy-to-disambiguate high-frequency
configurations.  The similarity-based methods perform up to 40\% better on this particular
task.
\end{abstract}
\section{Introduction}

Data sparseness is an inherent problem in statistical methods for
natural language processing.  Such methods use statistics on the
relative frequencies of configurations of elements in a training
corpus to learn how to evaluate alternative analyses or
interpretations of new samples of text or speech.  The most likely
analysis will be taken to be the one that contains the most frequent
configurations.  
The problem of data sparseness, also known as the
zero-frequency problem \cite{Witten+Bell:91a}, arises when analyses
contain configurations that never occurred in the training corpus.
Then it is not possible to estimate probabilities from observed
frequencies, and some other estimation scheme that can generalize from
the training data has to be used.

In language processing applications, the sparse data problem occurs
even for very large data sets.  For example,
\citeA{Essen:92a} report that in a 75\%-25\% split of the
million-word LOB corpus, 12\% of the bigrams in the test partition
did not occur in the training portion.  For trigrams, the sparse data
problem is even more severe: for instance, researchers at IBM
\cite{Brown:92c} examined a training corpus consisting of almost 366
million English words, and discovered that one can expect 14.7\% of
the word triples in any new English text to be absent from the
training sample.  Thus, estimating the probability of unseen
configurations is crucial to accurate language modeling, since
the aggregate probability of these unseen events can be significant.

We focus here on a particular kind of configuration, {\em word
cooccurrence}.  Examples of such cooccurrences include relationships
between head words in syntactic constructions (verb-object or
adjective-noun, for instance) and word sequences ($n$-grams).  In
commonly used models, the probability estimate for a previously unseen
cooccurrence is a function of the probability estimates for the words
in the cooccurrence.  For example, in word bigram models, the
probability $P(w_2 | w_1)$ of a {\em conditioned} word $w_2$ that has
never occurred in training following the {\em conditioning} word $w_1$
is typically calculated from the probability of $w_2$, as estimated by
$w_2$'s frequency in the corpus \cite{Jelinek:92a,Katz:87a}.  This
method makes an independence assumption on the cooccurrence of
$w_1$ and $w_2$: the more frequent $w_2$ is, the higher the
estimate of $P(w_2 | w_1)$ will be, regardless of $w_1$.

Class-based and similarity-based models provide an alternative to the
independence assumption. In these models, the relationship between
given words is modeled by analogy with other words that are in some
sense similar to the given ones.

For instance, \citeA{Brown:92c} suggest a class-based $n$-gram
model in which words with similar cooccurrence distributions are
clustered into word classes.  The cooccurrence probability of a given
pair of words is then estimated according to an averaged cooccurrence
probability of the two corresponding classes.  \citeA{Pereira:93a}
propose a ``soft'' distributional clustering scheme for certain
grammatical cooccurrences in which membership of a word in a class is
probabilistic.  Cooccurrence probabilities of words are then modeled
by averaged cooccurrence probabilities of word clusters.

\citeA{Dagan+Marcus+Markovitch:93a,Dagan+Marcus+Markovitch:95a}
present a similarity-based model, 
which avoids building clusters.  Instead,
each word is modeled by its own specific class, a  set of
words that are most similar to it.  Using this scheme, they predict
which unobserved cooccurrences are more likely than others.  Their
model, however, does not provide probability estimates and so cannot
be used as a component of a larger probabilistic model, as would be
required in, say, speech recognition.

Class-based and similarity-based methods for cooccurrence
modeling may at first sight seem to be special cases of clustering and
weighted nearest-neighbor approaches used widely in machine learning and
pattern recognition
\cite{Aha+al:91a,Cover+Hart:67a,Duda+Hart:73a,Stanfill+Waltz:86a,Devroye+al:96,Atkeson+Moore+Schaal:97a}.
There are important differences between those methods and ours.  Clustering and
nearest-neighbor techniques often rely on representing objects as
points in a multidimensional space with coordinates determined by the
values of intrinsic object features.  However, in most
language-modeling settings, all we know about a word are
the frequencies of its cooccurrences with other words in certain
configurations. Since the purpose of modeling is to estimate the
probabilities of cooccurrences, the same cooccurrence statistics are
the basis for both the similarity measure and the model predictions.
That is, the only means we have for measuring word similarity
are the predictions words make about what words they cooccur with,
whereas in typical instance or
(non-distributional) clustering learning methods, word similarity is
defined from intrinsic features independently of the predictions
(cooccurrence probabilities or classifications) associated with
particular words (see for instance the work of \citeA{Cardie:93a},
\citeA{Ng+Lee:96a}, \citeA{Ng:97b}, and
\citeA{Zavrel+Daelemans:97a}).

\subsection{Main Contributions}
\label{sec:maincontrib}
Our main contributions are a general scheme for using word similarity 
to improve the probability estimates of back-off models, and a 
comparative analysis of several similarity measures and parameter 
settings in two important language processing tasks, language modeling 
and disambiguation, showing that similarity-based estimates are indeed useful. 

In our initial study, a language-model evaluation, 
we used a similarity-based model
to estimate unseen bigram probabilities for {\em Wall Street Journal}
text and compared it to a standard back-off model \cite{Katz:87a}.
Testing on a held-out sample, the similarity model achieved a 20\%
perplexity reduction over back-off for unseen bigrams.  These
constituted 10.6\% of the test sample, leading to an overall reduction
in test-set perplexity of 2.4\%. The similarity-based model was also 
tested in a speech-recognition task, where it yielded
a statistically significant reduction
(32 versus 64 mistakes in cases where there was disagreement with
the back-off model) in recognition error.

In the disambiguation evaluation, we compared several variants of our
initial method and the {\em cooccurrence smoothing} method of
\citeA{Essen:92a} against the estimation method of Katz in a
decision task involving unseen pairs of direct objects and verbs.  We
found that all the similarity-based models performed almost 40\%
better than back-off, which  yielded about 49\% accuracy
in our experimental setting.  Furthermore, a scheme based on the
Jensen-Shannon divergence \cite{Rao:82a,Lin:91}\footnote{To the best
of our knowledge, this is the first use of this particular
distribution dissimilarity function in statistical language
processing.  The function itself is implicit in earlier work on
distributional clustering \cite{Pereira:93a} and has been used by
Tishby (p.c.)  in other distributional similarity work.
\citeA{Finch:thesis} discusses its use in word clustering, but does
not provide an experimental evaluation on actual data.}  yielded
statistically significant improvement in error rate over cooccurrence
smoothing.

We also investigated the effect of removing extremely low-frequency
events from the training set.  We found that, in contrast to back-off
smoothing, where such events are often discarded from training with
little dis\-cern\-ible effect, sim\-i\-lar\-ity-based smoothing methods suffer
noticeable performance degradation when singletons (events that occur
exactly once) are omitted.

The paper is organized as follows.  Section \ref{sec:models} describes
the general similarity-based framework; in particular, Section
\ref{sec:measures} presents the functions we use as measures of
similarity.  Section \ref{sec:langmod} details our initial language
modeling experiments.  Section \ref{sec:disamb} describes our
comparison experiments on a pseudo-word disambiguation task. Section
\ref{sec:otherwork} discusses related work. Finally, Section
\ref{sec:conclusion} summarizes our contributions and outlines future
directions.

\section{Distributional Similarity Models}
\label{sec:models}

We wish to model conditional probability distributions arising from
the cooccurrence of linguistic objects, typically words, in certain
configurations. We thus consider pairs $(w_{1},w_{2})\in V_{1}\times
V_{2}$ for appropriate sets $V_{1}$ and $V_{2}$, not necessarily
disjoint. In what follows, we use subscript $i$ for
the $i^{th}$ element of a pair; thus $P(w_{2}|w_{1})$ is the conditional
probability (or rather, some empirical estimate drawn from a  base
language model, the true probability
being unknown) that a pair has second element $w_{2}$ given that its
first element is $w_{1}$; and $P(w_1 | w_2)$ denotes the
probability estimate, according to the base language model, that $w_1$
is the first word of a pair given that the second word is $w_2$.
$P(w)$ denotes the base estimate for the unigram probability of word
$w$.

A similarity-based language model consists of three parts: a scheme
for deciding which word pairs require a similarity-based estimate, a
method for combining information from similar words, and, of course, a
function measuring the similarity between words.  We give the details
of each of these three parts in the following three sections.
We will only be concerned with similarity between words in
$V_1$, which are the conditioning events for the probabilities
$P(w_{2}|w_{1})$ that we want to estimate.

\subsection{Discounting and Redistribution}
\label{sec:redistribute}

Data sparseness makes the {\em maximum
likelihood estimate (MLE)} for word pair probabilities unreliable.
The MLE for the probability of a word pair $(w_1,w_2)$, conditional on
the appearance of word $w_1$,  is simply
\begin{equation}
P_{ML}(w_2|w_1) = \frac{c(w_1,w_2)}{c(w_1)},
\end{equation}
where $c(w_1,w_2)$ is the frequency of $(w_1,w_2)$ in the training
corpus and $c(w_1)$ is the frequency of $w_1$. However, $P_{ML}$ is
zero for any unseen word pair, that is, any such pair would be 
predicted as impossible. More generally, the MLE is unreliable for 
events with small nonzero counts as well as for those with zero counts. 
In the language modeling literature, the term {\em smoothing} is used 
to refer to methods for adjusting the probability estimates of 
small-count events away from the MLE to try to alleviate its 
unreliability. Our proposals address the zero-count problem 
exclusively, and we rely on existing techniques to smooth other small 
counts.

Previous proposals for the zero-count problem
\cite{Good:53a,Jelinek:92a,Katz:87a,Church+Gale:91a} adjust the MLE
so that the total probability of seen word pairs is less than one,
leaving some probability mass to be redistributed among the unseen
pairs. In general, the adjustment involves either {\em interpolation},
in which the MLE is used in linear combination with an estimator
guaranteed to be nonzero for unseen word pairs, or {\em discounting},
in which a reduced MLE is used for seen word pairs, with the
probability mass left over from this reduction used to model unseen
pairs.

The {\em back-off} method of \citeA{Katz:87a} is a prime example of 
discounting:
\begin{equation}
\hat{P}(w_2| w_1) = \left\{\begin{array}{l@{\hspace{2ex}}l}
P_d(w_2 | w_1) & c(w_1,w_2) > 0 \\
\alpha(w_1)P_r(w_2 | w_1) & c(w_1,w_2) = 0
\end{array}\right.\qquad ,\label{genmodel}
\end{equation}

\noindent where $P_d$ represents the Good-Turing discounted estimate
\cite{Katz:87a} for seen word pairs, and $P_r$ denotes the model for
probability redistribution among the unseen word pairs.  $\alpha(w_1)$
is a normalization factor.  Since an extensive comparison study by
\citeA{Chen+Goodman:96} indicated that back-off is better than
interpolation for estimating bigram probabilities, we will not
consider interpolation methods here; however, one could easily 
incorporate similarity-based estimates into an interpolation
framework as well.

In his original back-off model, Katz used $P(w_2)$ as the model for
predicting $\hat{P}(w_2| w_1)$ for unseen word pairs, that is, his 
model backed off to a unigram model for unseen bigrams. However, it is 
conceivable that backing off to a more detailed model than unigrams 
would be advantageous. Therefore, we generalize Katz's 
formulation by writing $P_r(w_2|w_1)$ instead of $P(w_2)$,
enabling us to use similarity-based estimates for unseen word pairs
instead of unigram frequency.  Observe that similarity
estimates are used for unseen word pairs only.

We next investigate estimates for $P_r(w_2| w_1)$ derived by averaging
information from words that are distributionally similar to $w_1$.

\subsection{Combining Evidence}
\label{sec:combine}
Similarity-based models make the following assumption: if word $w_1'$
is ``similar'' to word $w_1$, then $w_1'$ can yield information about
the probability of unseen word pairs involving $w_1$.  We use a
weighted average of the evidence provided by similar words, or {\em
neighbors}, where the
weight given to a particular word $w_1'$ depends on its similarity to
$w_1$.

More precisely, let $W(w_1,w_1')$ denote an increasing function of the
similarity between $w_1$ and $w_1'$, and let ${\cal S}(w_1)$
denote the set of words most similar to $w_1$.  Then the
general form of similarity model we consider is a $W$-weighted linear
combination of predictions of similar words:
\begin{equation}
P_{\smrm{SIM}}(w_2|w_1) = \sum_{w_1' \in {\cal S}(w_1)}
\frac{W(w_1,w_1')}{{\rm norm}(w_1)}{P(w_2|w_1')}\qquad ,
\label{sim-formula}
\end{equation}
where ${\rm norm}(w_1) = \sum_{w_1' \in {\cal S}(w_1)}W(w_1,w_1')$ is a
normalization factor.
According to this formula, $w_2$ is more likely to occur
with $w_1$ if it tends to occur with the words that are most similar to $w_1$.

Considerable latitude is allowed in defining the set ${\cal S}(w_1)$,
as is evidenced by previous work that can be put in the above form.
\citeA{Essen:92a} and \citeA{Karov:96a} (implicitly) set ${\cal S}(w_1) = V_1$.
However, it may be desirable to restrict ${\cal S}(w_1)$ in some
fashion for efficiency reasons, especially if $V_1$ is large.  For instance, in the
language modeling application of Section \ref{sec:langmod}, we use the closest $k$
or fewer words $w_1'$ such that the dissimilarity between $w_1$ and
$w_1'$ is less than a threshold value $t$; $k$ and $t$ are tuned
experimentally.

One can directly replace $P_r(w_2|w_1)$ in the back-off equation
(\ref{genmodel}) with $P_{\smrm{SIM}}(w_2|w_1)$.  However, other
variations are possible, such as interpolating with the unigram
probability $P(w_2)$:
$$P_r(w_2|w_1) = \gamma P(w_2) + (1 - \gamma)
P_{\smrm{SIM}}(w_2|w_1)\qquad .$$ This represents, in effect, a linear
combination of the similarity estimate and the back-off estimate: if
$\gamma = 1$, then we have exactly Katz's back-off scheme.  In the
language modeling task (Section \ref{sec:langmod}) we set $\gamma$
experimentally; to simplify our comparison of different similarity
models for sense disambiguation (Section \ref{sec:disamb}), we set
$\gamma$ to $0$.

It would be possible to make $\gamma$ 
depend on $w_1$, so that the contribution of the similarity estimate 
could vary among words.  Such dependences are often used in 
interpolated models 
\cite{Jelinek:80a,Jelinek:92a,Saul+Pereira-97:factored} and are indeed 
advantageous.  However, since they introduce hidden variables, they 
require a more complex training algorithm, and we did not pursue that 
direction in the present work.

\subsection{Measures of Similarity}
\label{sec:measures}

We now consider several word similarity measures that can be derived
automatically from the statistics of a training corpus, as opposed to
being derived from manually-constructed word classes
\cite{Yarowsky:92b,Resnik:92a,Resnik:95a,Luk:95a,Lin:97a}.  Sections
\ref{sec:KL} and \ref{sec:avg} discuss two related
information-theoretic functions, the KL divergence and the
Jensen-Shannon divergence.  Section \ref{sec:Lone} describes the
$L_1$ norm, a geometric distance function.  Section \ref{sec:conf}
examines the confusion probability, which has been previously employed
in language modeling tasks.  There are, of course, many other possible
functions;  we have opted to restrict our attention to this reasonably
diverse set.

For each function, a corresponding weight function $W(w_1,w_1')$ is
given.  The choice of weight function is to some extent arbitrary; the
requirement that it be increasing in the similarity between $w_1$ and
$w_1'$ is not extremely constraining.  While clearly performance
depends on using a good weight function, it would be impossible to
try all conceivable $W(w_1,w_1')$.  Therefore, in section
\ref{sec:wsd-eval}, we describe experiments evaluating
similarity-based models both with and without weight functions.

All the similarity functions we describe depend on some base language
model $P(w_{2} | w_{1})$, which may or may not be the Katz discounted
model $\hat{P}(w_{2} | w_{1})$ from Section \ref{sec:redistribute}
above.  While we discuss the complexity of computing each similarity
function, it should be noted that in our current implementation, this
is a one-time cost: we construct the $|V_1| \times |V_1|$ matrix of
word-to-word similarities before any parameter training takes place.

\subsubsection{KL divergence}
\label{sec:KL}
The {\it Kullback-Leibler (KL) divergence} is a
standard infor\-mation-theoretic measure of the dissimilarity between two
probability mass functions \cite{Kullback:59,Cover:91a}.  We can apply it
to the
conditional distributions induced by words in $V_1$ on words
in $V_2$:
\begin{equation}
D(w_1 \Vert w_1') = \sum_{w_2} P(w_2|w_1) \log 
\frac{P(w_2|w_1)}{P(w_2|w_1')}\qquad .
\label{KL}
\end{equation}
$D(w_1 \Vert w_1')$ is non-negative, and is zero if and only if $P(w_{2}|w_1)
= P(w_{2} | w_1')$ for all $w_{2}$. However, the KL divergence is non-symmetric and
does not obey the triangle inequality.

For $D(w_1\Vert w_1')$ to be defined it must be the case that
$P(w_2|w_1') > 0$ whenever $P(w_2|w_1) > 0$. Unfortunately, this 
generally does not hold for MLEs based on samples; we must use smoothed
estimates that redistribute some probability mass to zero-frequency
events. But this forces the sum in (\ref{KL}) to be over all $w_2\in
V_2$, which makes this calculation expensive for large vocabularies.

Once the divergence $D(w_1 \Vert w_1')$ is computed, we set
\[
W_D(w_1,w_1')= 10^{-\beta D(w_1||w_1')} \qquad .
\] 
The role of the free parameter $\beta$ is to control the relative
influence of the neighbors closest to $w_1$: if $\beta$ is high, then
$W_D(w_1,w_1')$ is non-negligible only for those $w_1'$ that are
extremely close to $w_1$, whereas if $\beta$ is low, distant neighbors
also contribute to the estimate.
We chose a negative exponential function of the KL divergence for 
the weight function by analogy with the form of the cluster 
membership function in related distributional clustering work \cite{Pereira:93a} 
and also because that is the form for the probability that 
$w_{1}$'s distribution arose from a sample drawn from the 
distribution of $w_{1}'$ \cite{Cover:91a,Lee:thesis}. However, these reasons are heuristic 
rather than theoretical, since we do not have a rigorous probabilistic
justification for similarity-based methods.

\subsubsection{Jensen-Shannon divergence}
\label{sec:avg}
A related measure is the {\em Jensen-Shannon
divergence} \cite{Rao:82a,Lin:91}, which can be defined as the average of
the KL divergence of each of two distributions to their average  distribution:
\begin{equation}
\JS(w_1,w_1') = \frac{1}{2}\left[ D \left( w_1 \biggl\Vert \frac{w_1 +
w_1'}{2} \right) + D \left(
w_1' \biggl\Vert
\frac{w_1 + w_1'}{2} \right)\right]\qquad,
\label{avg}
\end{equation}
where $(w_1 + w_1')/2$ is shorthand for the
distribution
\[
\frac{1}{2}(P(w_{2}|w_1) + P(w_{2}|w_1'))\qquad. \]
Since
the KL divergence is nonnegative, $\JS(w_1, w_1')$ is also nonnegative. Furthermore,
letting $p(w_{2}) = P(w_{2}|w_{1})$ and $p'(w_{2}) = P(w_{2}|w_{1}')$,
it is easy to see that
\begin{equation}
\JS(w_{1}, w_{1}')= H\left(\frac{p+p'}{2}\right) - \frac{1}{2}H(p) -
\frac{1}{2}H(p')\qquad,
\label{eqn:jensen-entropy}
\end{equation}
where $H(q) = - \sum_{w} q(w)\log q(w)$ is the entropy of the
discrete density $q$. This equation shows that $\JS$ gives the
information gain achieved by distinguishing the two distributions $p$
and $p'$ (conditioning on contexts $w_{1}$ and $w_{1}'$) over pooling
the two distributions (ignoring the distinction between $w_{1}$ and
$w_{1}'$).

It is also easy to see that $\JS$ can be computed efficiently,
since it depends only on those conditioned words that occur in both
contexts. Indeed, letting ${\cal C}=\{w_{2}:p(w_{2})>0,p'(w_{2})>0\}$,
and grouping the terms of (\ref{eqn:jensen-entropy}) appropriately, we
obtain
\[
\JS(w_1,w_1')   =  \log 2 +  \frac{1}{2}\sum_{w_{2}\in {\cal C}}
\left\{ h\left(\p + \q\right)
- h(\p) - h(\q) \right\}\qquad , \]
where $h(x) = -x \log x$.
Therefore,  $\JS(w_1,w_1')$ is bounded, ranging between $0$ and $
\log 2$, and smoothed estimates are not required because probability
ratios are not involved.

As in the KL divergence case, we set $W_{\JS}(w_1,w_1')= 10^{-\beta
\JS(w_1,w_1')}$; $\beta$ plays the same role as before.

\subsubsection{$L_1$ norm}
\label{sec:Lone}
The {\it $L_1$ norm} is defined as
\begin{equation}
L(w_1,w_1') = \sum_{w_2} \left| P(w_2|w_1) - P(w_2|w_1') 
\right|\qquad .
\label{var}
\end{equation}
By grouping terms as before, we can
express $L(w_1,w_1')$ in a form depending only on
the ``common'' $w_2$:
\[
L(w_1, w_1') =  2 - \sum_{w_2\in {\cal C}} \p  - \sum_{w_2\in {\cal
C}} \q + \sum_{w_2\in {\cal C}} |\p - \q |\qquad .
\]
It follows from  the triangle inequality 
that $0 \leq L(w_1, w_1') \leq 2$, with equality to $0$ if and only if
there are no words $w_2$ such that both $P(w_2| w_1)$ and
$P(w_2|w_1')$ are strictly positive.

Since we require a weighting scheme that is decreasing in $L$, we
set
\[ W_L(w_1,w_1') = (2
- L(w_1,w_1'))^\beta\qquad,\] with $\beta$ again free.\footnote{We
experimented with using $10^{-\beta L(w_1,w_1')}$ as well, but it
yielded poorer performance results.}  As before, the higher $\beta$ is,
the more relative influence is accorded to the nearest neighbors.

It is interesting to note the following relations between the $L_1$
norm, the KL-divergence, and the Jensen-Shannon divergence.
\citeA{Cover:91a} give the following lower bound:
\[
L(w_1, w_1') \leq  \sqrt{ D(w_1 || w_1') \cdot 2 \log b}\qquad,
\label{Lone-bound}
\]
where $b$ is the base of the logarithm function.  \citeA{Lin:91}
notes that $L$ is an upper bound for $J$:
$$\JS(w_{1}, w_{1}') \leq L(w_1,w_1')\qquad .$$

\subsubsection{Confusion probability}
\label{sec:conf}

Extending work by \citeA{Sugawara+al:85a}, \citeA{Essen:92a}
used {\it confusion probability} to estimate word cooccurrence
probabilities.\footnote{Actually, they present two alternative
definitions.  We use their model 2-B, which they found yielded the
best experimental results.}  They report 14\% improvement in test-set
perplexity (defined below) on a small corpus.  The confusion probability
was also used by \citeA{Grishman+Sterling:93a} to estimate the
likelihood of selectional patterns.

The confusion probability is an estimate of the probability that word
$w_1'$ can be substituted for word $w_1$, in the sense of being found
in the same contexts:

\begin{eqnarray*}
P_{\smrm{C}}(w_1' | w_1) &= &  \sum_{w_2} \frac{P(w_1
| w_2) P(w_1' | w_2) P(w_2)} {P(w_1)} \\
& = & W_{\smrm{C}}(w_1,w_1') \nonumber
\end{eqnarray*}
\noindent ($P(w_1)$ serves as a normalization factor).  In
contrast to the distance functions described above, $\conf$ has the
curious property that $w_1$ may not necessarily be the ``closest''
word to itself, that is, there may exist a word $w_1'$ such that
$\confarg > P_{\smrm{C}}(w_1|w_1)$; see Section \ref{sec:sampleclose}
for an example.

The confusion probability can be computed from empirical estimates
provided all unigram estimates are nonzero (as we assume throughout).
In fact, the use of smoothed estimates such as those provided by
Katz's back-off scheme is problematic, because those estimates
typically do not preserve consistency with respect to marginal
estimates and Bayes's rule
(that is, it may be that $\sum_{w_2} P(w_1|w_2)P(w_2) \neq
P(w_1)$). However, using consistent estimates (such as the MLE), we
can safely apply Bayes's rule to rewrite $\conf$ as follows:
\begin{equation}
\confarg = \sum_{w_2} \frac{P(w_2|w_1)}{P(w_2)} \cdot
P(w_2| w_1') P(w_1')\qquad .
\label{confuse-alternate}
\end{equation}

\noindent As with the Jensen-Shannon divergence and the $L_1$ norm,
this sum  requires computation only over the ``common'' $w_2$'s.

Examination of Equation (\ref{confuse-alternate}) reveals an important
difference between the confusion probability and the functions $D$,
$\JS$, and $L$ described in the previous sections.  Those functions
rate $w_1'$ as similar to $w_1$ if, roughly, $P(w_2|w_1')$ is high
when $P(w_2 | w_1)$ is.  $\confarg$, however, is greater for those
$w_1'$ for which $P(w_1', w_2)$ is large when $P(w_2|w_1)/P(w_2)$ is.
When this ratio is large, we may think of $w_2$ as
being exceptional, since if $w_2$ is infrequent, we do not expect
$P(w_2|w_1)$ to be large.

\subsubsection{Summary}

Several features of the measures of similarity listed above are
summarized in Table \ref{table:simsum}. ``Base LM constraints'' are
conditions that must be satisfied by the probability estimates of the
base language model.  The last column indicates whether the weight
$W(w_1, w_1')$ associated with each similarity function depends on a
parameter that needs to be tuned experimentally.

\begin{table*}[t]
\begin{center}
\begin{tabular}{l|l|c|c}
name &  range &  base LM constraints & tune? \\ \hline
$D$ & $[0, \infty]$ & $P(w_2 | w_1') \neq 0$ if $P(w_2|w_1) \neq 0$ &
yes \\
$\JS$ & $[0,\log 2]$ &none & yes\\
$L$ & $[0,2]$ & none & yes\\
$P_{\smrm{C}}$ & $[0, \frac{1}{2} \max_{w_2} P(w_2)]$ & Bayes
consistency & no\\
\end{tabular}
\end{center}
\caption{\label{table:simsum} Summary of Similarity Function
Properties }
\end{table*}

\section{Language Modeling}
\label{sec:langmod}

The goal of our first set of experiments, described in this section,
was to provide proof of concept by showing that similarity-based
models can achieve better language modeling performance than back-off.
We therefore only used one similarity measure.  The success of these
experiments convinced us that similarity-based methods are worth
examining more closely; the results of our second set of experiments,
comparing several similarity functions on a pseudo-word disambiguation
task, are described in the next section.

Our language modeling experiments used a similarity-based model, with
the KL divergence as (dis)similarity measure, as an alternative to
unigram frequency when backing off in a bigram model.  That is, we
used the bigram language model defined by:
\begin{eqnarray}
\hat{P}(w_2| w_1) &= &\left\{
\begin{array}{l@{\hspace{2ex}}l}
P_d(w_2 | w_1) & c(w_1,w_2) > 0 \\
\alpha(w_1)P_r(w_2 | w_1) & c(w_1,w_2) = 0
\end{array}\right.\nonumber \\
P_r(w_2|w_1) & = &\gamma P(w_2) + (1 - \gamma) P_{\smrm{SIM}}(w_2|w_1)
\nonumber \\
P_{\smrm{SIM}}(w_2|w_1) & = &\sum_{w_1' \in {\cal S}(w_1)}
\frac{W(w_1,w_1')}{{\rm norm}(w_1)}{P(w_2|w_1')} \label{eqn:psim}\\
W(w_1,w_1') &= &  10^{-\beta D(w_1||w_1')} \nonumber\qquad,
\end{eqnarray}
where $V_1 = V_2 = V$, the entire vocabulary.
As noted earlier, the estimates of $P(w_{2}|w_{1}')$ must be smoothed
to avoid division by zero when computing $D(w_{1}||w_{1}')$; we employed
the standard Katz bigram back-off model for that purpose.  Since
$|V| = 20,000$ in this application, we considered only a small fraction
of $V$ in computing $P_{\smrm{SIM}}$, using the tunable thresholds
$k$ and $t$ described in Section \ref{sec:combine} for this purpose.

The standard evaluation metric for language models is the
likelihood of the test data according to the model, or, more
intuitively, the test-set {\em perplexity}
\[ \sqrt[N]{\prod_{i=1}^{N}P(w_i|w_{i-1})^{-1}} \qquad,\]
which represents the average number of alternatives presented by the
(bigram) model after each test word. Thus, a better model will have a lower
perplexity. In our task, lower perplexity will indicate better
prediction of unseen bigrams.

We evaluated the above model by comparing its test-set perplexity and
effect on speech-recognition accuracy with the baseline bigram
back-off model developed by MIT Lincoln Laboratories for the {\em Wall
Street Journal} (WSJ) text and dictation corpora provided by ARPA's
HLT program \cite{Paul:91}.%
\footnote{The ARPA WSJ development corpora come in two versions, one
with verbalized punctuation and the other without. We used the latter
in all our experiments.} The baseline back-off model follows the Katz
design, except that, for the sake of compactness, all frequency one bigrams are
ignored. The counts used in this model and in ours were obtained from
40.5 million words of WSJ text from the years 1987-89.

For perplexity evaluation, we tuned the similarity model parameters by
minimizing perplexity on an additional sample of 57.5 thousand words
of WSJ text, drawn from the ARPA HLT development test set.  The best
parameter values found were $k=60$, $t=2.5$, $\beta=4$ and
$\gamma=0.15$. For these values, the improvement in perplexity for
unseen bigrams in a held-out 18 thousand word sample (the ARPA HLT 
evaluation test set) is just over
20\%. Since unseen bigrams comprise 10.6\% of this sample, the
improvement on unseen bigrams corresponds to an overall test set
perplexity improvement of 2.4\% (from 237.4 to 231.7). 
\begin{table*}
\begin{center}
\begin{tabular}{rr@{.}lr@{.}lr@{.}lr@{.}lr@{.}l}
$k$ & \multicolumn{2}{c}{$t$} & \multicolumn{2}{c}{$\beta$} &
\multicolumn{2}{c}{$\gamma$} & \multicolumn{2}{c}{training reduction
(\%)}& \multicolumn{2}{c}{test reduction
(\%)}\\
\hline
60 & 2&5 & 4&0 & 0&15 & 18&4 & 20&51 \\
50 & 2&5 & 4&0 & 0&15 & 18&38 & 20&45 \\
40 & 2&5 & 4&0 & 0&2 & 18&34 & 20&03 \\
30 & 2&5 & 4&0 & 0&25 & 18&33 & 19&76 \\
70 & 2&5 & 4&0 & 0&1 & 18&3 & 20&53 \\
80 & 2&5 & 4&5 & 0&1 & 18&25 & 20&55 \\
100 & 2&5 & 4&5 & 0&1 & 18&23 & 20&54 \\
90 & 2&5 & 4&5 & 0&1 & 18&23 & 20&59 \\
20 & 1&5 & 4&0 & 0&3 & 18&04 & 18&7 \\
10 & 1&5 & 3&5 & 0&3 & 16&64 & 16&94
\end{tabular}
\end{center}
\caption{Perplexity Reduction on Unseen Bigrams for Different Model Parameters}
\label{perp-results}
\end{table*}
Table \ref{perp-results} shows reductions in training and test
perplexity, sorted by training reduction, for different choices of the
number $k$ of closest neighbors used. The values of $\beta$, $\gamma$
and $t$ are the best ones found for each $k$.%
\footnote{Values of
$\beta$ and $t$ refer to base 10 logarithms and exponentials in all
calculations.}

From equation (\ref{eqn:psim}), it is clear that the computational cost of
applying the similarity model to an unseen bigram is
$O(k)$. Therefore, lower values for $k$ (and
$t$) are computationally preferable.
\todo{How many times is the $t$ limit actually invoked?}
From the table, we can see that reducing $k$ to 30 incurs
a penalty of less than 1\% in the perplexity improvement, so
relatively low values of $k$ appear to be sufficient to achieve most
of the benefit of the similarity model. As the table also shows, the
best value of $\gamma$ increases as $k$ decreases; that is, for lower
$k$, a greater weight is given to the conditioned word's frequency.
This suggests that the predictive power of neighbors beyond the
closest 30 or so can be modeled fairly well by the overall frequency
of the conditioned word.

The bigram similarity model was also tested as a language model in
speech recognition. The test data for this experiment were pruned word
lattices for 403 WSJ closed-vocabulary test sentences.  Arc scores in these
lattices are sums of an acoustic score (negative log likelihood) and a
language-model score, which in this case was the negative log probability
provided by the baseline bigram model.

From the given lattices, we constructed new lattices in which the arc
scores were modified to use the similarity model instead of the
baseline model.  We compared the best sentence hypothesis in each
original lattice with the best hypothesis in the modified one, and
counted the word disagreements in which one of the hypotheses was
correct.  There were a total of 96 such disagreements; the similarity
model was correct in 64 cases, and the back-off model in 32. This
advantage for the similarity model is statistically significant at the
0.01 level. The overall reduction in error rate is small, from 21.4\%
to 20.9\%, because the number of disagreements is small compared with
the overall number of errors in the recognition setup employed in
these experiments.

Table \ref{rec-eg} shows some examples of speech recognition
disagreements between the two models. The hypotheses are labeled `B'
for back-off and `S' for similarity, and the bold-face words are
errors. The similarity model seems to be  better at modeling
regularities such as semantic parallelism in lists and avoiding a past
tense form after ``to.'' On the other hand, the similarity model
makes several mistakes in which a function word is inserted in a place
where punctuation would be found in written text.

\begin{table*}
\begin{center}
\begin{tabular}{l|l}
B & commitments \ldots from leaders {\bf felt the} three point six
billion dollars \\
\hline S & commitments \ldots from leaders fell to three point six
billion dollars \\
\hline \hline B & followed by France the US {\bf
agreed in}  Italy  \\
\hline S & followed by France the US Greece
\ldots Italy \\
\hline\hline B & he whispers to {\bf made a} \\
\hline
S & he whispers to an aide \\
\hline\hline B & the necessity for change
{\bf exist} \\
 \hline S & the necessity for change exists \\
\hline\hline\hline B & without \ldots additional reserves Centrust  would
have reported \\
\hline S & without \ldots additional reserves {\bf
of} Centrust would have reported \\
\hline \hline  B & in the darkness
past the church \\
\hline S & in the darkness {\bf passed} the church
\end{tabular}
\end{center}
\caption{Speech Recognition Disagreements between Models}
\label{rec-eg}
\end{table*}

\section{Word-Sense Disambiguation}
\label{sec:disamb}

Since the experiments described in the previous section demonstrated
promising results for similarity-based estimation, we ran a second set
of experiments designed to help us compare and analyze the somewhat
diverse set of similarity measures given in Table \ref{table:simsum}.
Unfortunately, the KL divergence and the confusion probability have
different requirements on the base language model, and so we could not
run a direct four-way comparison.  As explained below, we elected to
omit the KL divergence from consideration.

We chose to evaluate the three remaining measures on a word sense
disambiguation task, in which each method was presented with a noun
and two verbs, and was asked which verb was more likely to have the
noun as a direct object.  Thus, we did not measure the absolute
quality of the assignment of probabilities, as would be the case in a
perplexity evaluation, but rather the relative quality.  We could
therefore ignore constant factors, which is why we did not normalize
the similarity measures.

\subsection{Task Definition}

In the usual word sense disambiguation problem, the method to be
tested is presented with an ambiguous word in some context, and is
asked to identify the correct sense of the word from that context.
For example, a test instance might be the sentence fragment ``robbed
the bank''; the question is whether ``bank'' refers to a river bank, a
savings bank, or perhaps some other alternative meaning.

While sense disambiguation is clearly an important problem for
language processing applications, as an
evaluation task it presents numerous experimental difficulties.
First, the very notion of ``sense'' is not clearly defined; for
instance, dictionaries may provide sense distinctions that are too
fine or too coarse for the data at hand.  Also, one needs to have
training data for which the correct senses have been assigned;
acquiring these correct senses generally requires considerable human
effort.  Furthermore, some words have many possible senses, whereas
others are essentially monosemous; this means that test cases are not
all uniformly hard.

To circumvent these and other difficulties, we set up a pseudo-word
disambiguation experiment \cite{Schutze92a,Gale92b}, the format of
which is as follows.  First, a list of {\em pseudo-words} is
constructed, each of which is the combination of two different words
in $V_2$.  Each word in $V_2$ contributes to exactly one pseudo-word.
Then, every $w_2$ in the test set is replaced with its corresponding
pseudo-word.  For example, if a pseudo-word is created out of
the words ``make'' and ``take'', then the data is altered as follows:
\begin{center}
\begin{tabular}{llcl}
make & plans & $\Rightarrow$ & \{make, take\} plans \\
take & action &  $\Rightarrow$ & \{make, take\} action \\
\end{tabular}
\end{center}
The method being tested must choose between the two
words that make up the pseudo-word.  

The advantages of using pseudo-words are two-fold.  First, the
alternative ``senses'' are under the control of the experimenter.
Each test instance presents exactly two alternatives to the
disambiguation method, and the alternatives can be chosen to be of the
same frequency, the same part of speech, and so on.  Secondly, the
pre-transformation data yields the correct answer, so that no
hand-tagging of the word senses is necessary.  These advantages make
pseudo-word experiments an elegant and simple means to test the
efficacy of different language models; of course they may not provide
a completely accurate picture of how the models would perform in
real disambiguation tasks, 
although one could create more realistic settings by making
pseudo-words out of more than two words, varying the frequencies of
the alternative pseudo-senses, and so on.  

For ease of comparison, we did not consider interpolation with unigram
probabilities.  Thus, the model we used for these
experiments differs slightly from that used in the language modeling
tests; it can be summarized as follows:

\begin{eqnarray*}
\hat{P}(w_2| w_1) &= &\left\{
\begin{array}{l@{\hspace{2ex}}l}
P_d(w_2 | w_1) & c(w_1,w_2) > 0 \\
\alpha(w_1)P_r(w_2 | w_1) & c(w_1,w_2) = 0
\end{array}\right.\nonumber \\
P_r(w_2|w_1) & = & P_{\smrm{SIM}}(w_2|w_1)
\nonumber \\
P_{\smrm{SIM}}(w_2|w_1) & = &\sum_{w_1' \in {\cal S}(w_1)}
\frac{W(w_1,w_1')}{{\rm norm}(w_1)}{P(w_2|w_1')} \nonumber \\
\end{eqnarray*}

\subsection{Data}

We used a statistical part-of-speech tagger \cite{Church:88a} and
pattern matching and concordancing tools (due to David Yarowsky) to
identify transitive main verbs ($V_2$) and head nouns ($V_1$) of the
corresponding direct objects in 44 million words of 1988 Associated
Press newswire.  We selected the noun-verb pairs for the $1000$ most
frequent nouns in the corpus.  These pairs are undoubtedly somewhat
noisy given the errors inherent in the part-of-speech tagging and
pattern matching. 

We used $80\%$, or $587,833$, of the pairs so derived for building 
models, reserving $20\%$ for testing purposes.  As some, but not all, 
of the similarity measures require smoothed models, we calculated both 
a Katz back-off model ($P = \hat{P}$ in equation (\ref{genmodel}), 
with $P_r(w_2|w_1) = P(w_2)$), and a maximum-likelihood model ($P = 
P_{\smrm{ML}}$).  Furthermore, we wished to evaluate the hypothesis 
that a more compact language model can be built without affecting 
model quality by deleting {\em singletons}, word pairs that occur only 
once, from the training set.  This claim had been made in particular 
for language modeling \cite{Katz:87a}.
We therefore built four base models, summarized in Table
\ref{table:langmod}.

\begin{table}[ht]
\begin{center}
\begin{tabular}{c|cc}
 	& with singletons 	& no singletons
 \\
 & (587,833 pairs)	&(505,426 pairs) \\ \hline
MLE	&  \MLEone\			& \MLEoone\	\\
Katz 	& \boone			& \booone\			\\
\end{tabular}
\end{center}
\caption{\label{table:langmod} Base Language Models}
\end{table}

Since we wished to test the effectiveness of using similarity for
unseen word cooccurrences, we removed from the test data any
verb-object pairs that occurred in the training set; this resulted in
$17,152$ {\em unseen} pairs (some occurred multiple times).  The unseen
pairs were further divided into five equal-sized parts, $T_1$ through
$T_5$, which formed the basis for fivefold cross-validation: in each
of five runs, one of the $T_i$ was used as a performance test set,
with the other four  combined into one set used for tuning parameters
(if necessary) via a simple grid search that evaluated the 
error on the tuning set at regularly spaced points in parameter 
space.  Finally, test pseudo-words
were created from pairs of verbs with similar frequencies, so as to
control for word frequency in the decision task.  
Our method was to simply rank the verbs by frequency and create
pseudo-words out of all adjacent pairs (thus, each verb participated in
exactly one pseudoword).  Table \ref{table:pseudowords} lists some
randomly chosen pseudowords and the frequencies of the corresponding verbs.

\begin{table}[ht]
\begin{center}
\begin{tabular}{@{\hspace{1.5in}}l@{\hspace{1.5in}}}
make (14782)/take (12871)\\
fetch (35)/renegotiate (35) \\
magnify (13)/exit (13) \\
meeet (1)/stupefy (1) \\
relabel (1)/entomb (1) 
\end{tabular}
\end{center}
\caption{\label{table:pseudowords} Sample pseudoword verbs and
frequencies.  The word ``meeet'' is a typo occurring in the corpus.}
\end{table}

We use error rate as
our performance metric, defined as
\[
\frac{1}{N} (\mbox{\# of incorrect choices } + (\mbox{\# of
ties})/2)
\]
where $N$ was the size of the test corpus.  A tie occurs
when the two words making up a pseudo-word are deemed equally likely.

\subsection{Baseline Experiments}

The performances of the four base language models are shown in
Table \ref{table:baseline}.  \MLEone\ and \MLEoone\ both have
error rates of exactly $.5$ because the test sets consist of unseen
bigrams, which are all assigned a probability of $0$ by maximum-likelihood
estimates, and thus are all ties for this method.  The back-off models
\boone\ and \booone\ also perform similarly.

\begin{table}[ht]
\begin{center}
\begin{tabular}{l|lllll}
 & $T_1$ & $T_2$ & $T_3$ & $T_4$ & $T_5$ \\ \hline
\MLEone\ & .5 & .5 & .5 & .5 & .5 \\
\MLEoone\ & \H{ } & \H{ } & \H{ } & \H{ } & \H{ } \\
\boone\ & 0.517 & 0.520 & 0.512 & 0.513 &0.516 \\
\booone\ & 0.517 & 0.520 & 0.512 & 0.513 &0.516 \\
\end{tabular}
\end{center}
\caption{\label{table:baseline} Base Language Model Error Rates}
\end{table}

Since the back-off models consistently performed worse than the MLE
models, we chose to use only the MLE models in our subsequent
experiments.  Therefore, we only ran comparisons between the measures
that could utilize unsmoothed data, namely, the $L_1$ norm,
$L(w_1,w_1')$; the Jensen-Shannon divergence, $\JS(w_1, w_1')$; and the
confusion probability, $P_{\smrm{C}}(w_1' | w_1)$. \footnote{It should
be noted, however, that on \boone\ data, the KL-divergence performed slightly
better than the $L_1$ norm.}  

\subsection{Sample Closest Words}
\label{sec:sampleclose}

In this section, we examine the closest words to a randomly selected
noun, ``guy'', according to the three measures $\Lone$, $\JS$, and $\conf$.

Table \ref{table:closeMLE1} shows the ten closest words, in order,
when the base language model is \MLEone.  There is some overlap between
the closest words for $\Lone$ and the closest words for $\JS$, but very
little overlap between the closest words for these measures and the
closest words with respect to $\conf$: only the words ``man'' and
``lot'' are common to all three.  Also observe that the word ``guy'' itself is
only  fourth on the list of words with the
highest confusion probability with respect to ``guy''.

\begin{table}[t]
\begin{center}
\begin{tabular}{lr@{.}l|lr@{.}l|lr@{.}l}
\multicolumn{3}{c}{$L$} &\multicolumn{3}{c}{$\JS$}
&\multicolumn{3}{c}{$\conf$} \\ \hline
GUY & 0&0  & GUY & 0&0  & role & 0&033 \\
kid & 1&23   & kid & 0&15   & people & 0&024 \\
lot & 1&35   & thing & 0&1645 & fire & 0&013 \\
thing & 1&39 & lot & 0&165   & GUY & 0&0127 \\
man & 1&46   & man & 0&175   & man & 0&012 \\
doctor & 1&46 & mother & 0&184 & year & 0&01 \\
girl & 1&48  & doctor & 0&185 & lot & 0&0095 \\
rest & 1&485 & friend & 0&186 & today & 0&009 \\
son & 1&497   & boy & 0&187 & way & 0&008778\\
bit & 1&498   & son & 0&188 & part & 0&008772 \\
\multicolumn{3}{c|}{(role: rank 173)}	&
\multicolumn{3}{c|}{(role:  rank 43)} &
\multicolumn{3}{c}{(kid:  rank 80)} \\
\end{tabular}
\end{center}
\caption{\label{table:closeMLE1} 10
closest words to the word ``guy''
for $\Lone$, $\JS$, and $\conf$, using \MLEone\ as the base language model.  The
rank of the words ``role'' and ``kid'' are also shown if they are not
among the top ten.}
\end{table}

Let us examine the case of the nouns ``kid'' and ``role'' more
closely.  According to the similarity functions $\Lone$ and $\JS$, ``kid'' is
the second closest word to ``guy'', and ``role'' is considered
relatively distant.  In the $\conf$ case, however, ``role'' has
the highest confusion probability with respect to ``guy,'' whereas
``kid'' has only the 80th highest confusion probability.  What
accounts for these differences?

Table \ref{table:verbs}, which gives the ten verbs most likely to
occur with ``guy'', ``kid'', and ``role'', indicates that both $\Lone$
and $\JS$ rate words as similar if they tend to cooccur with the same
verbs.  Observe that four of the ten most likely verbs to
occur with ``kid'' are also very likely to occur with ``guy'', whereas
only the verb ``play'' commonly occurs with both ``role'' and ``guy''.

\begin{table}[t]
\begin{center}
\begin{tabular}{l|l}
Noun & Most Likely Verbs \\ \hline
guy &  see get play let  give catch tell do pick need \\ \hline
kid & {\bf get} {\bf see} take help want {\bf tell} teach send {\bf give}
love \\
role & {\bf play} take lead support assume star expand accept sing limit \\
\end{tabular}
\end{center}
\caption{\label{table:verbs} For each
noun $w_1$, the ten verbs $w_2$ with
highest
$P(w_2|w_1)$.  Bold-face verbs occur with both the given noun and with
``guy.'' The base language model is \MLEone.}
\end{table}

\begin{table}
\begin{center}
\begin{tabular}{l}
(1) electrocute
(2) shortchange
(3) bedevil
(4) admire
(5) bore
(6) fool \\
$\quad$ (7) bless
$\cdots$
(26) play
$\cdots$
 (49) get
$\cdots$
\end{tabular}
\end{center}
\caption{\label{table:sigverbs}  Verbs with highest
$P(w_2|
\mbox{``guy''})/P(w_2)$ ratios.  The numbers in parentheses are ranks.}
\end{table}
If we sort the verbs by decreasing $P(w_2|
\mbox{``guy''})/P(w_2)$, a different order emerges (Table
\ref{table:sigverbs}):
``play'', the most likely verb to cooccur with ``role'', is
ranked higher than ``get'', the most likely verb to cooccur with ``kid'',
thus indicating why ``role'' has a higher confusion probability with
respect to ``guy'' than ``kid'' does.

\begin{table}
\begin{center}
\begin{tabular}{lr@{.}l|lr@{.}l|lr@{.}l}
\multicolumn{3}{c}{$L$} &\multicolumn{3}{c}{$\JS$}
&\multicolumn{3}{c}{$\conf$} \\ \hline
GUY & 0&0   & GUY & 0&0    & role & 0&05   \\
kid & 1&17    & kid & 0&15    & people & 0&025  \\
lot & 1&40   & thing & 0&16   & fire & 0&021  \\
thing & 1&41  & lot & 0&17     & GUY & 0&018   \\
reason & 1&417  & mother & 0&182  & work & 0&016    \\
break & 1&42   & answer & 0&1832  & man & 0&012    \\
ball & 1&439   & reason & 0&1836  & lot & 0&0113   \\
answer & 1&44  & doctor & 0&187  & job & 0&01099     \\
tape & 1&449    & boost & 0&189   & thing & 0&01092   \\
rest & 1&453    & ball & 0&19    & reporter & 0&0106\\
\end{tabular}
\end{center}
\caption{\label{table:closeMLEo1} 10
closest words to the word ``guy'' for
$\Lone$, $\JS$, and
$\conf$, using \MLEoone\ as the base language model.}
\end{table}

Finally, we examine the effect of deleting singletons from the base
language model.  Table \ref{table:closeMLEo1} shows the ten closest
words, in order, when the base language model is \MLEoone. The relative
order of the four closest words remains the same; however, the next
six words are quite different from those for \MLEone.  This data
suggests that the effect of singletons on calculations of similarity
is quite strong, as is borne out by the experimental evaluations
described in Section \ref{sec:wsd-eval}. 

We conjecture that this effect is due to the fact that there are many
very low-frequency verbs in the data (65\% of the verbs appeared with
10 or fewer nouns; the most common verb occurred with 710
nouns). Omitting singletons involving such verbs may well drastically
alter the number of verbs that cooccur with both of two given nouns
$w_1$ and $w_1'$. Since the similarity functions we consider in
this set of experiments depend on such words, it is not surprising
that the effect of deleting singletons is rather dramatic. In
contrast, a back-off language model is not as sensitive to missing
singletons because of the Good-Turing discounting of small counts and
inflation of zero counts.

\subsection{Performance of Similarity-Based Methods}
\label{sec:wsd-eval}
\begin{figure}[htb]
\epsfscaledbox{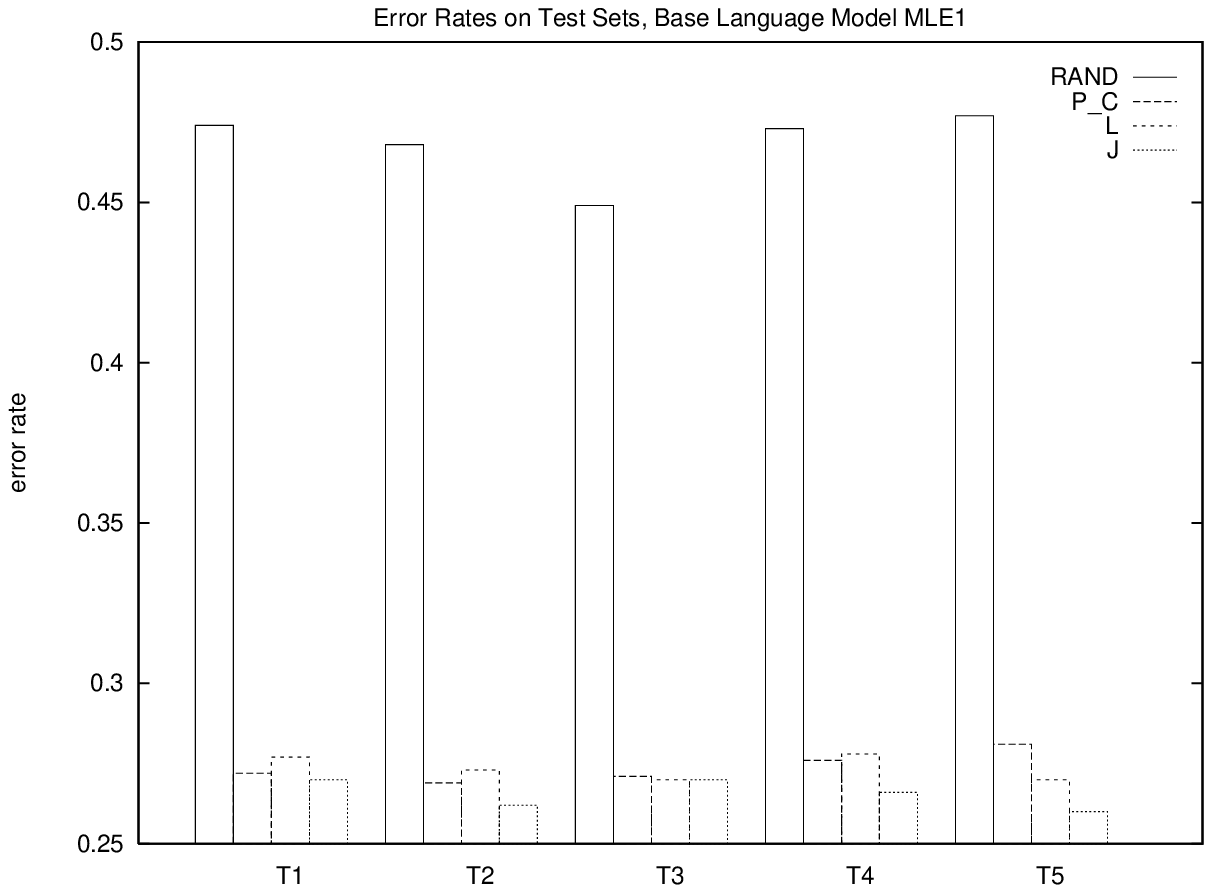}{4in}
\caption{\label{fig:MLE1}  Error rates for each test set, where the
base language model was \MLEone.  The methods, going from left to right,
are $\rand\;$, $\conf$, $L$, and $\JS$. The performances shown
are for settings of $\beta$ that were optimal for the corresponding
training set.  $\beta$ ranged from $4.0$ to
$4.5$ for $L$ and from $20$ to $26$ for $\JS$.}
\end{figure}

Figure \ref{fig:MLE1} shows the results of our experiments on the five
test sets, using \MLEone\ as the base language model.  The parameter
$\beta$ was always set to the optimal value for the corresponding
training set.  \rand, which is shown for comparison purposes, simply
chooses the weights $W(w_1, w_1')$ randomly.  ${\cal S}(w_1)$ was set
equal to $V_1$ in all cases.

The sim\-ilar\-ity-based methods consistently outperformed 
Katz's back-off meth\-od and the MLE (recall that both yielded error rates of
about .5) by a large margin, indicating that information from other
word pairs is very useful for unseen pairs when unigram frequency is
not informative.  The similarity-based methods also do much better
than $\rand$, which indicates that it is not enough to simply combine
information from other words arbitrarily: word similarity should be
taken into account.  In all cases, $\JS$ edged out the other methods.
The average improvement in using $\JS$ instead of $\conf$ is .0082;
this difference is significant to the .1 level ($p < .085$), according
to the paired t-test.

\begin{figure}[htb]
\epsfscaledbox{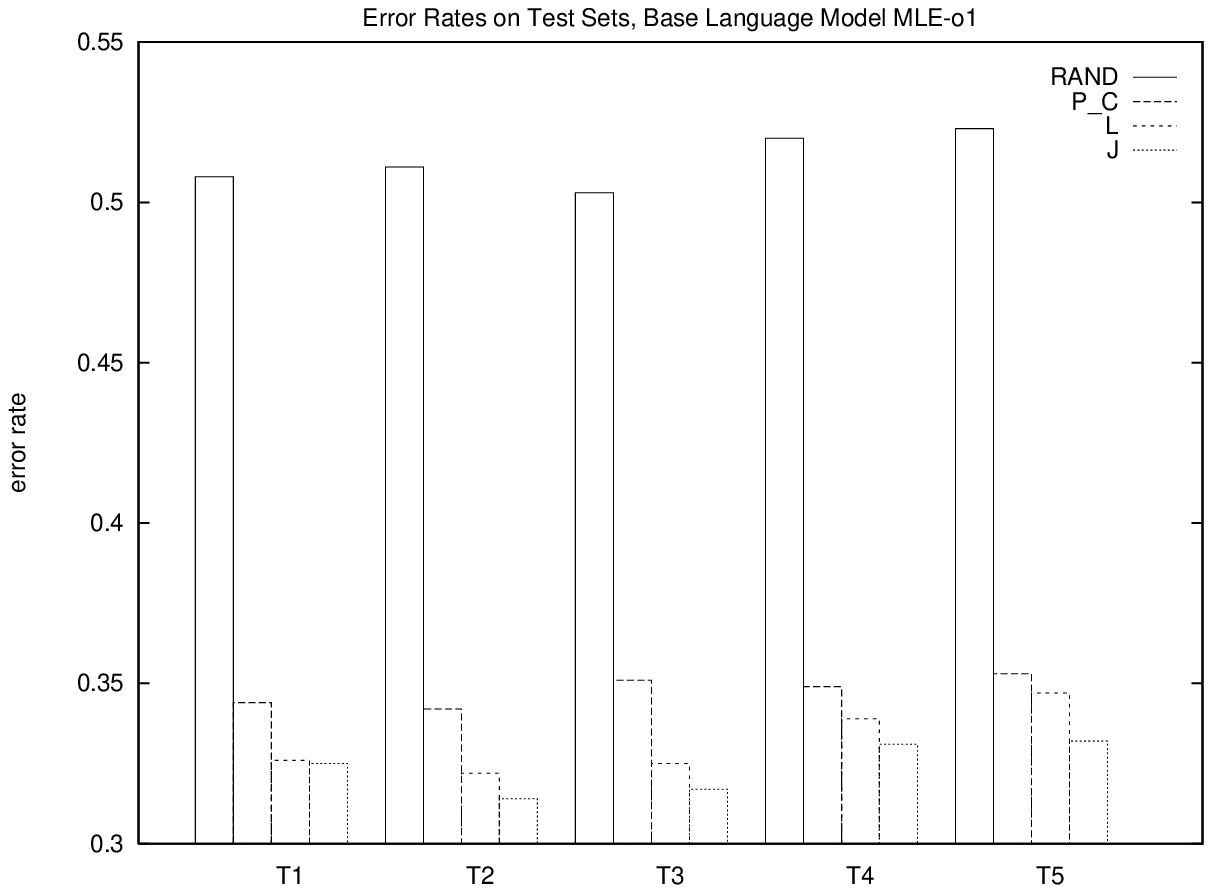}{4in}
\caption{\label{fig:MLEo1}  Error rates for each test set, where the
base language model was \MLEoone.  $\beta$ ranged from $6$ to
$11$ for $L$ and from $21$ to $22$ for $\JS$.}
\end{figure}

The results for the \MLEoone\ case are depicted in Figure
\ref{fig:MLEo1}.  Again, we see the similarity-based methods achieving
far lower error rates than the MLE, back-off, and $\rand$ methods, and
again, $\JS$ always performed the best.  However, omitting singletons
amplified the disparity between $\JS$ and $\conf$: the average
difference was $.024$, which is significant to the .01 level (paired
t-test).  

An important observation is that all methods, including \rand,
suffered a performance hit if singletons were deleted from the base
language model.  This seems to indicate that seen bigrams should be
treated differently from unseen bigrams, even if the seen bigrams are
extremely rare.  We thus conclude that one cannot create a compressed
similarity-based language model by omitting singletons without hurting
performance, at least for this task.

We now analyze the role of the parameter $\beta$.  Recall that $\beta$
appears in the weight functions for the 
Jensen-Shannon divergence and the $L_1$ norm:
\[ W_{\JS}(w_1,w_1')= 10^{-\beta\JS(w_1,w_1')} ~~,~~
W_L(w_1,w_1') = (2 - L(w_1,w_1'))^\beta\qquad.\]
It controls the relative influence of the most similar words: 
their influence increases with higher values of $\beta$.

\begin{figure}[htb]
\epsfscaledbox{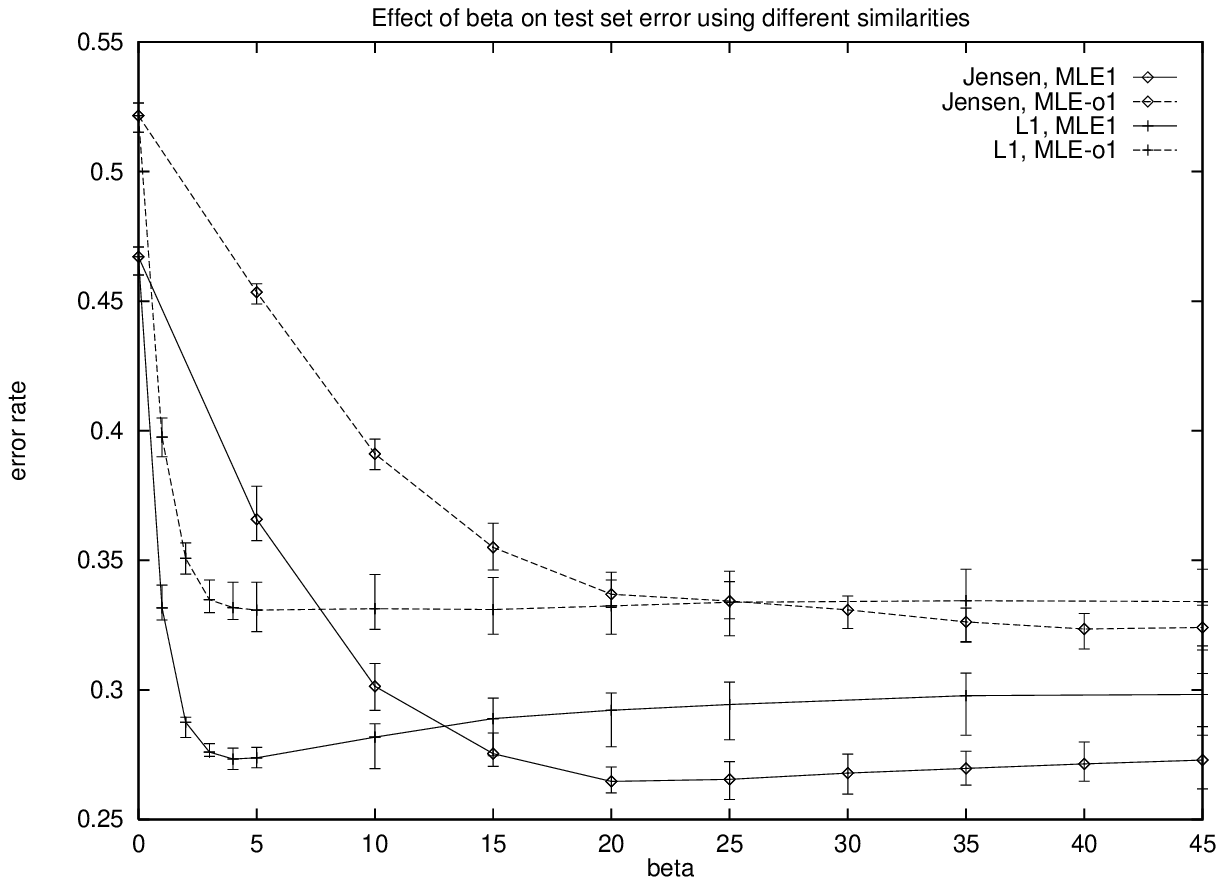}{4in}
\caption{\label{fig:betas} Average and range of test-set error rates
as $\beta$ is varied.  The similarity function is indicated by the
point style; the base language model is indicated by the line style.}
\end{figure}

Figure \ref{fig:betas} shows how the value of $\beta$ affects
disambiguation performance.  Four curves are shown, each corresponding
to a choice of similarity function and base language model.  The
error bars depict the average and range of error rates over the five
disjoint test sets.

It is immediately clear that to get good performance results,
$\beta$ must be set much higher for the Jensen-Shannon divergence
than for the $L_1$ norm.  This phenomenon results from the fact that
the range of possible values for $\JS$ is much smaller than that for
$\Lone$.  This ``compression'' of $\JS$ values requires a large
$\beta$ to scale differences of distances correctly.

We also observe that setting $\beta$ too low causes substantially
worse error rates; however, the curves level off rather than moving
upwards again.  That is, as long as a sufficiently large value is
chosen, setting $\beta$ suboptimally does not greatly impact
performance.  Furthermore, the shape of the curves is the same for
both base language models, suggesting that the relation between
$\beta$ and test-set performance is relatively insensitive to
variations in training data.

\begin{figure}[htb]
\epsfscaledbox{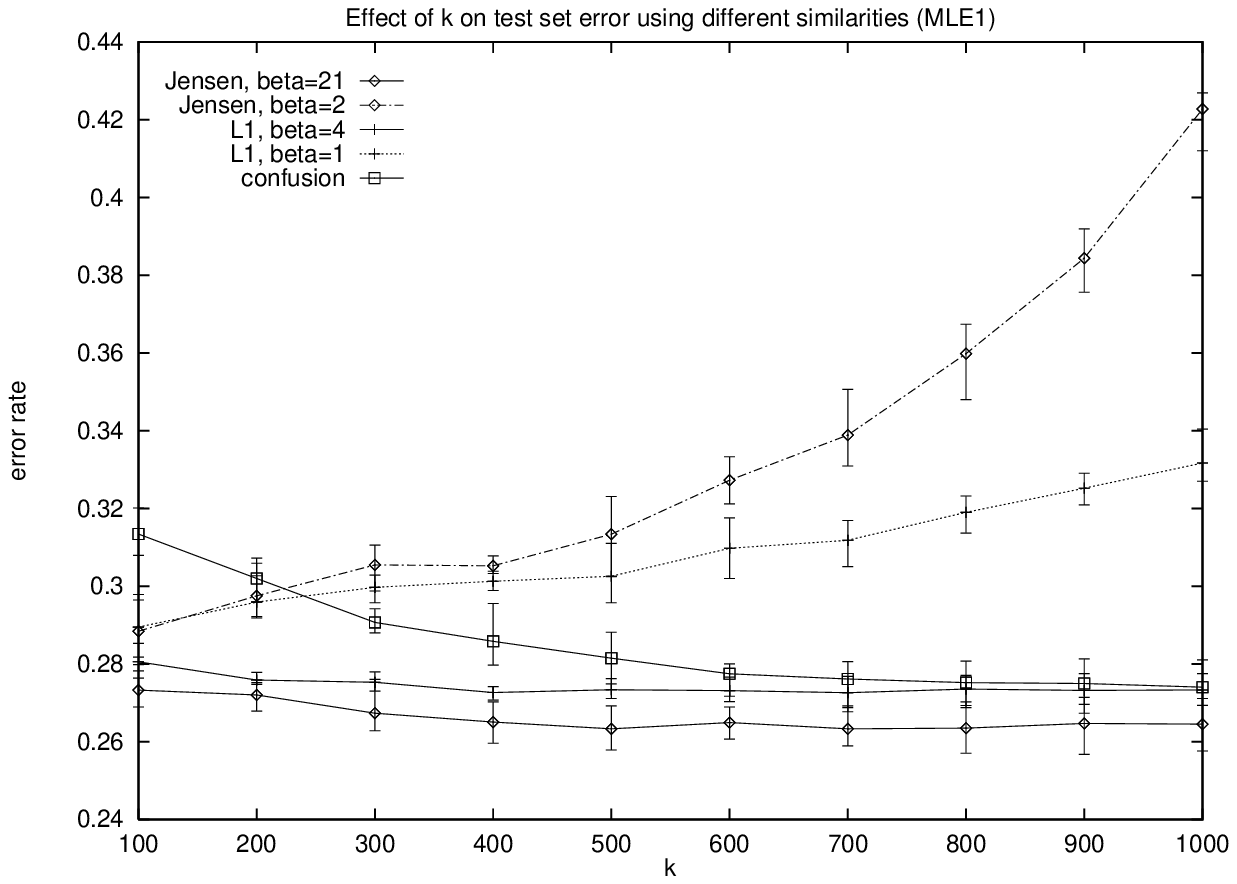}{4in}
\caption{\label{fig:varyk-betas} Average and range of test-set error
rates as $k$ is varied.  The base language model was \MLEone.  
The similarity function is indicated by the point style; the dashed
and dotted lines indicate a suboptimal choice of $\beta$.}
\end{figure}

The fact that higher values of $\beta$ seem to lead to better error
rates suggests that $\beta$'s role is to filter out distant neighbors.
To test this hypothesis, we experimented with using only the $k$ most
similar neighbors.  Figure \ref{fig:varyk-betas} shows how the error
rate depends on $k$ for different fixed values of $\beta$.  The two
lowest curves depict the performance of the Jensen-Shannon divergence
and the $L_1$ norm when $\beta$ is set to the optimal value with
respect to average test set performance; it appears that the more
distant neighbors have essentially no effect on error rate because their
contribution to the sum (\ref{eqn:psim})
is negligible.  In contrast, when too low a value of $\beta$ is chosen
(the upper two curves), distant neighbors are weighted too
heavily.  In this case, including more distant neighbors causes
serious degradation of performance.

Interestingly, the behavior of the confusion probability is different
from these two cases: adding more neighbors actually improves the
error rate.  This seems to indicate that the confusion probability is
not correctly ranking similar words in order of informativeness.
However, an alternative explanation is that $\conf$ is at a
disadvantage only because it is not being employed in the context of a
tunable weighting scheme.  

\begin{figure}[htb]
\epsfscaledbox{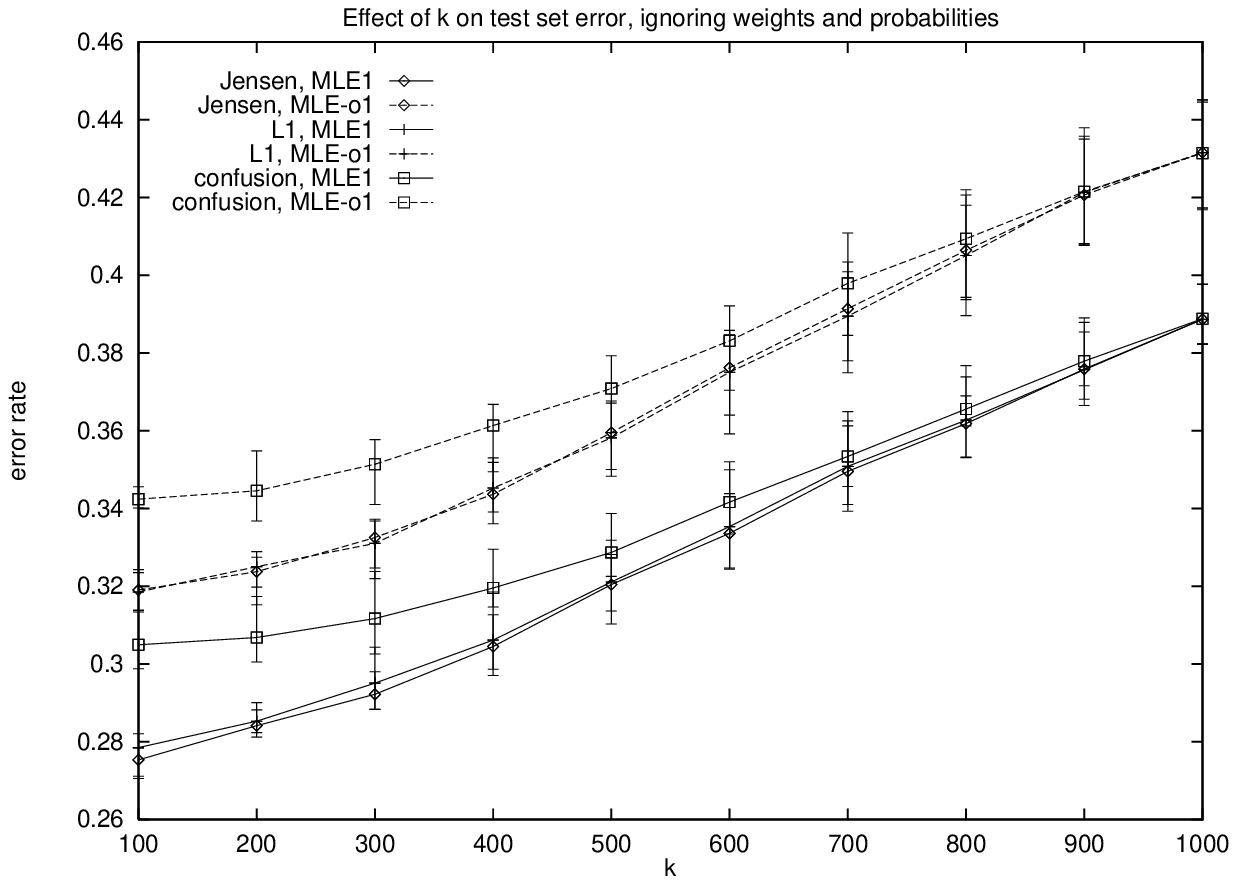}{4in}
\caption{\label{fig:varyk-noweights} Average and range of
voting-scheme test-set error rates as $k$ is varied.  The similarity
function is indicated by the point style; the base language model is
indicated by the line style.}
\end{figure}

To distinguish between these two possibilities, we
ran an experiment that dispensed with weights altogether.  Instead, we
took a vote of the $k$ most similar neighbors: the alternative chosen
as more likely was the one preferred by a majority of the most similar
neighbors (note that we  ignored the
{\em degree} to which alternatives were preferred).  The results are
shown in Figure \ref{fig:varyk-noweights}.

We see that the $k$ most similar neighbors according to $\JS$ and
$\Lone$ were always more informative than those chosen according to the
confusion probability, with the largest performance gaps occurring for
low $k$ (of course, all methods performed the same for $k = 1000$,
since in that case they were using the same set of neighbors).  This
graph provides clear evidence that the confusion probability is not
as good a measure of the informativeness of other words.

\section{Related Work}
\label{sec:otherwork}
There is a large body of work on notions of work similarity, word 
clustering, and their applications. It is impossible to compare all 
those methods directly, since the assumptions, experimental settings 
and applications of methods vary widely. Therefore, the discussion 
below is mainly descriptive, highlighting some of the main
similarities and differences between the methods. 

\subsection{Statistical similarity and clustering for disambiguation
and language modeling} 
Our work is an instance of a growing body of research on using word 
similarity to improve performance in language-processing problems.  
Similarity-based algorithms either use the similarity scores 
between a word and other words directly in making their predictions, 
or rely on similarity scores between a word and representatives of 
precomputed similarity classes.

An early attempt to automatically classify words into semantic classes
was carried out in the Linguistic String Project
\cite{Grishman86}. Semantic classes were derived from similar
cooccurrence patterns of words within syntactic relations.
Cooccurrence statistics were then considered at the class level and
used to alleviate data sparseness in syntactic disambiguation.

\citeA{Schutze92,Schutze93} captures contextual word 
similarity by first reducing the dimensionality of a context 
representation using singular value decomposition and then using 
the reduced-dimensionality representation to characterize the possible 
contexts of a word.  This information is used for word sense 
disambiguation.  All occurrences of an ambiguous word are clustered 
and each cluster is mapped manually to one of the senses of the word.  
The context vector of a new occurrence of the ambiguous word is mapped 
to the nearest cluster which determines the sense for that occurrence.  
Sch\"{u}tze emphasizes that his method avoids clustering words into a 
pre-defined set of classes, claiming that such clustering is likely to 
introduce artificial boundaries that cut off words from part of their 
semantic neighborhood.

\citeA{Karov:96a} have also addressed the data 
sparseness problem in word sense disambiguation by using word 
similarity.  They use a circular definition for both a word similarity 
measure and a context similarity measure.  The circularity is resolved 
by an iterative process in which the system learns a set of typical 
usages for each of the senses of an ambiguous word.  Given a new 
occurrence of the ambiguous word the system selects the sense whose 
typical context is most similar to the current context, applying a 
procedure which resembles the sense selection process of Sh\"{u}tze.

Our scheme for employing word similarity in disambiguation was 
influenced by the work of 
\citeA{Dagan+Marcus+Markovitch:93a,Dagan+Marcus+Markovitch:95a}.  
Their method computes a word similarity measure directly from word 
cooccurrence data.  A word is then modeled by a set of most similar 
words, and the plausibility of an unseen cooccurrence is judged by the 
cooccurrence statistics of the words in this set.  The similarity 
measure is a weighted Tanimoto measure, a version of which was also 
used by \citeA{Grefenstette92,Grefenstette94}.  Word association is 
measured by mutual information, following earlier work on word 
similarity by \citeA{Hindle90}.

The method of 
\citeauthor{Dagan+Marcus+Markovitch:93a,Dagan+Marcus+Markovitch:95a} 
does not provide probabilistic models.  Disambiguation decisions are 
based on comparing scores for different alternatives, but they do not 
produce explicit probability estimates and therefore cannot be 
integrated directly within a larger probabilistic framework.  The 
cooccurrence smoothing model of \citeA{Essen:92a}, like our model, 
produces explicit estimates of word cooccurrence probabilities based 
on the cooccurrence statistics of similar words.  The similarity-based 
estimates are interpolated with direct estimates of $n$-gram 
probabilities to form a smoothed $n$-gram language model.  Word 
similarity in this model is computed by the confusion probability 
measure, which we described and evaluated earlier.

Several language modeling methods produce similarity-based probability 
estimates through class-based models.  These methods do not use a 
direct measure of the similarity between a word and other words, but 
instead cluster the words into classes using a global optimization 
criterion.  \citeA{Brown:92c} present a class-based {\em n}-gram 
model which records probabilities of sequences of word classes instead 
of sequences of individual words.  The probability estimate for a 
bigram which contains a particular word is affected by bigram 
statistics for other words in the same class, where all words in the 
same class are considered similar in their cooccurrence behavior.  
Word classes are formed by a bottom-up hard-clustering algorithm whose 
objective function is the average mutual information of class 
cooccurrence.  
\citeA{Ushioda96} introduces several improvements to 
mutual-information clustering.  His method, which was applied to 
part-of-speech tagging, records all classes which contained a 
particular word during the bottom-up merging process.  The word is 
then represented by a mixture of these classes rather than by a single 
class.  

The algorithms of \citeA{Kneser+Ney:93a} and \citeA{Ueberla:94a}
are similar to that of \citeA{Brown:92c}, although a different
optimization criterion is used, and the number of clusters remains
constant throughout the membership assignment process.
\citeA{Pereira:93a} use a formalism from statistical mechanics to
derive a top-down soft-clustering algorithm with probabilistic class
membership.  Word cooccurrence probability is then modeled by a
weighted average of class cooccurrence probabilities, where the
weights correspond to membership probabilities of words within
classes.

\subsection{Thesaurus-based similarity} 

The approaches described in the previous section induce word 
similarity relationships or word clusters from cooccurrence statistics 
in a corpus.  Other researchers developed methods which quantify 
similarity relationships based on information in the manually crafted 
WordNet thesaurus \cite{Miller90}.  \citeA{Resnik:92a,Resnik:95a} 
proposes a node-based approach for measuring the similarity between a 
pair of words in the thesaurus and applies it to various 
disambiguation tasks.  His similarity function is an 
information-theoretic measure of the informativeness of the least 
general common ancestor of the two words in the thesaurus 
classification.  \citeA{Jiang97} combine the node-based approach with 
an edge-based approach, where the similarity of nodes in the thesaurus 
is influenced by the path that connects them.  Their similarity method 
was tested on a data set of word pair similarity ratings derived from 
human judgments.

\citeA{Lin:97a,Lin:98} derives a general concept-similarity
measure from assumptions on desired properties of similarity.  His
measure is a function of the number of bits required to describe each of
the two concepts as well as their ``commonality''.  He then describes
an instantiation of the measure for a hierarchical thesaurus and
applies it to WordNet as part of a word sense disambiguation
algorithm.

\subsection{Contextual similarity for information retrieval} 

Query expansion in information retrieval (IR) provides an additional 
motivation for automatic identification of word similarity.  One line 
of work in the IR literature considers two words as similar if they 
occur often in the same documents.  Another line of work considers the 
same type of word similarity we are concerned with, that is, 
similarity measured derived from word-cooccurrence statistics.

\citeA{Grefenstette92,Grefenstette94} argues that 
cooccurrence within a document yields similarity judgements that 
are not sharp enough for query expansion.  Instead, he extracts coarse 
syntactic relationships from texts and represents a word by the set of 
its word-cooccurrences within each relation.  Word similarity is 
defined by a weighted version of the Tanimoto measure which compares 
the cooccurrence statistics of two words.  The similarity method was 
evaluated by measuring its impact on retrieval performance.

\citeA{Ruge92} also extracted word cooccurrences within
syntactic relationships and evaluated several similarity measures on
those data, focusing on versions of the cosine measure. The similarity
rankings obtained by these measures were compared to those produced by
human judges.

\section{Conclusions}
\label{sec:conclusion}
Similarity-based language models provide an appealing approach for 
dealing with data sparseness.  In this work, we proposed a general 
method for using similarity-based models to improve the estimates of 
existing language models, and we evaluated a range of similarity-based 
models and parameter settings on important language-processing tasks.  
In the pilot study, we compared the language modeling performance of a 
similarity-based model with a standard back-off model.  While the 
improvement we achieved over a bigram back-off model is statistically 
significant, it is relatively modest in its overall effect because of 
the small proportion of unseen events.  In a second, more detailed 
study we compared several similarity-based models and parameter 
settings on a smaller, more manageable word-sense disambiguation task.  
We observed that the similarity-based methods perform much better on 
unseen word pairs, with the measure based on the Jensen-Shannon 
divergence being the best overall.

Our experiments were restricted to bigram probability estimation for 
reasons of simplicity and computational cost. However, the relatively 
small proportion of unseen bigrams in test data makes the effect of 
similarity-based methods necessarily modest in the overall tasks. We 
believe that the benefits of similarity-based methods would be more 
substantial in tasks with a larger proportion of unseen events, for 
instance language modeling with longer contexts.
There is no obstacle in principle to doing this: in the trigram case,
for example, we would still be determining the probability of pairs
$V_{1} \times V_{2}$, but $V_1$ would consist of word pairs instead of
single words.  However, the number of possible similar events to a
given element in $V_{1}$ is then much larger than in the bigram case. Direct
tabulation of the events most similar to each event would thus not be practical, so more
compact or approximate representations would have to be investigated. It would also be
worth investigating the benefit of similarity-based methods to improve estimates for
low-frequency seen events. However, we would need to replace the back-off model by
another one that combines multiple estimates for the same event, for
example an interpolated model with context-dependent interpolation parameters.

Another area for further investigation is the relationship between 
similarity-based and class-based approaches. As mentioned in the
introduction, both rely on a common intuition, namely, that events can
be modeled to some extent by similar events.  Class-based methods are 
more computationally expensive at training time than nearest 
neighbor methods because they require searching for the best model 
structure (number of classes and, for hard clustering, class 
membership) and estimation of hidden parameters (class membership 
probabilities in soft clustering). On the other hand, class-based 
methods reduce dimensionality and are thus smaller and more efficient 
at test time. Dimensionality reduction has also been claimed to 
improve generalization to test data, but the evidence for this is 
mixed. Furthermore, some class-based models have theoretically 
satisfying probabilistic interpretations 
\cite{Saul+Pereira-97:factored}, whereas the justification for our 
similarity-based models is heuristic and empirical at present.
Given the variety of class-based
language modeling algorithms, as described in the section on related
work above, it is beyond the scope of this paper to compare the
performance of the two approaches. However, such a comparison, 
especially one that would bring both approaches under a common 
probabilistic interpretation, would be well worth pursuing.

\section*{Acknowledgments}
We thank Hiyan Alshawi, Joshua Goodman, Rebecca Hwa, Slava Katz, Doug 
McIlroy, Stuart Shieber, and Yoram Singer for many helpful 
discussions, Doug Paul for help with his bigram back-off model, and 
Andrej Ljolje and Michael Riley for providing word lattices for our 
speech recognition evaluation.  We also thank the reviewers of this 
paper for their constructive criticisms, and the editors of the present 
issue, Claire Cardie and Ray Mooney, for their help and suggestions.  
Portions of this work have appeared previously 
\cite{Dagan:94a,Dagan+Lee+Pereira:comp}; we thank the reviewers of 
those papers for their comments.  Part of this work was done while the 
first author was a member of technical staff and then a visitor at 
AT\&T Labs, and the second author was a graduate student at Harvard 
University and a summer visitor at AT\&T Labs.  The second author 
received partial support from the National Science Foundation under 
Grant No.~IRI-9350192, a National Science Foundation Graduate 
Fellowship, and an AT\&T GRPW/ALFP grant.

\bibliographystyle{apacite}
\bibliography{sim}
\end{document}